\documentclass{article}
\usepackage{arxiv}
\usepackage[utf8]{inputenc} 
\usepackage[T1]{fontenc}    
\usepackage{hyperref}       
\usepackage{url}            
\usepackage{booktabs}       
\usepackage{amsfonts}       
\usepackage{nicefrac}       
\usepackage{microtype}      
\usepackage{lipsum}
\usepackage{fancyhdr}       
\usepackage{graphicx}       
\usepackage{caption}
\usepackage{subcaption}

\graphicspath{{media/}}     
\usepackage[table,xcdraw]{xcolor}
\usepackage{pgf-pie}
\title{5G Long-Term and Large-Scale Mobile Traffic Forecasting} 

\author{
  Ufuk Uyan, M. Tuğberk İşyapar, Mahiye Uluyağmur Öztürk \\
  Huawei Turkey R\&D Center\\
  Istanbul\\
  \texttt{\{ufuk.uyan1, tugberk.isyapar1, mahiye.uluyagmur.ozturk\}@huawei.com} \\
}

\begin{document}
\maketitle

\begin{abstract}

\end{abstract}
It is crucial for the service provider to comprehend and forecast mobile traffic in large-scale cellular networks in order to govern and manage mechanisms for base station placement, load balancing, and network planning. The purpose of this article is to extract and simulate traffic patterns from more than 14,000 cells that have been installed in different metropolitan areas. To do this, we create, implement, and assess a method in which cells are first categorized by their point of interest and then clustered based on the temporal distribution of cells in each region. The proposed model has been tested using real-world 5G mobile traffic datasets collected over 31 weeks in various cities. We found that our proposed model performed well in predicting mobile traffic patterns up to 2 weeks in advance. Our model outperformed the base model in most areas of interest and generally achieved up to 15\% less prediction error compared to the naïve approach. This indicates that our approach is effective in predicting mobile traffic patterns in large-scale cellular networks.


\section{Introduction}
5G is the fifth generation of mobile network technology, and it promises faster speeds, lower latency, and improved capacity compared to its predecessors. A number of factors, such as the ongoing development of more intelligent mobile phones, the introduction of machine-to-machine connections, and the availability of enticing and data-intensive applications, are driving up the demand for mobile data traffic globally. Effective and precise mobile traffic forecasting is particularly important for 5G networks, which are expected to have much higher levels of traffic compared to previous generations of mobile networks.It has been well known that implementing traffic prediction can improve energy efficiency, ease resource allocation, provide the best user experience, and finally enable intelligent cellular networks. Traffic prediction has emerged as one of the main enabling technologies for autonomous networks, which is supported by the whole telecommunication sector, with the large-scale commercial deployment of the 5G network. Additionally, traffic forecasting is a crucial component of numerous transportation services, including navigation, route planning, and traffic control. By dynamically allocating network resources in accordance with actual traffic demand, precise short-term prediction of future traffic load information improves network energy efficiency, while long-term forecasting is crucial for network planning and base station localization.

For many practical applications, such as predicting the demand for mobile data traffic, time-series prediction techniques are essential. Generally speaking, there are two types of data prediction models: traditional and machine learning models\cite{Do}. Traditional techniques include statistical methods such as Autoregressive Integrated Moving Average (ARIMA) and its extensions, such as Seasonal ARIMA (SARIMA).
Due to numerous aspects, such as user mobility, the arrival pattern, and distinct user requirements, the pattern of network traffic is actually highly complex. To tackle these complexities, recent developments in machine learning and artificial intelligence have led to the development of more advanced time series prediction models. These models are able to learn and capture the underlying patterns and dependencies in the data and make more accurate predictions at a large scale. Recent developments in machine learning algorithms\cite{Chang} have established themselves as strong challengers to traditional statistical models in traffic prediction, capturing the intricate and nonlinear dependencies concealed in wireless traffic data.

In this paper, we conduct a study to comprehend and predict the mobile traffic of large-scale cellular data networks using a time series approach motivated by the aforementioned issues. In order to fully characterize the external factors influencing mobile network traffic volume, this document actively collects two types of cross-domain statistics, namely cell level traffic information and distribution of point of interest (POI). To aid in the estimation of mobile network traffic, correlations between these data sets and various types of cellular traffic are thoroughly investigated and studied. In order to model the diversity of patterns of various functional regions, city-wide cellular traffic data are then clustered into a number of groupings. Following clustering, the N-Beats\cite{N-BEATS} architecture estimates the downlink volume and average user count. Our main
technical contributions are summarized as follows:

\begin{itemize}
\item An algorithm for clustering cells with various temporal distribution characteristics into distinct clusters has been presented, in addition to capturing the pattern variety and similarity of cellular traffic of several city functional sectors.
\item A new method for cell-level and long-term traffic prediction has been proposed from data with a low sampling rate by taking advantage of the powerful capacity of the Neural basis expansion analysis for interpretable time series forecasting architecture (N-Beats).
\end{itemize}

The rest of this paper is structured as follows. Section II gives an overview of the related work. The cellular traffic data set is described in detail in Section III, along with a preliminary analysis relating to the data set's key findings. The general prediction framework built on the N-Beats architecture is then introduced in Section IV. In Section V, a thorough examination of the obtained results and findings are presented. Section VI brings the work to a close.

\section{Related Work}
Cellular traffic forecasting is inherently a time series forecasting issue. Existing works can broadly be grouped into two categories based on solving techniques, namely statistics-based methods and machine learning-based methods. For the first category, statistical or probability distributions are used to model and forecast cellular traffic. Auto-Regressive Integrated Moving Average (ARIMA), for example, is a linear time series model that captures trends and short-term dependency in traffic demand. Seasonality is included in more sophisticated models like Seasonal ARIMA (SARIMA)\cite{shu,yu} and exponential smoothing (like Holt-Winters)\cite{Tikunov, Bastos}. A Holt-Winters exponential smoothing technique is presented by Tikunov and Nishumura for short-term forecasting in cellular networks based on historical data\cite{Tikunov}. Similar to this, Shanghai's 9,000 base station data traffic is predicted using ARIMA\cite{Xu}. While neglecting significant spatial correlations, these approaches estimate user traffic at individual base stations\cite{Li}. To function successfully, they must have lengthy ongoing observations and solely take into account prior temporal information. Exploratory Factor Analysis\cite{Furno} was used to analyze complex spatiotemporal patterns of mobile data in order to profile network activity and identify land use.

Data-driven machine learning-based traffic prediction methods have emerged as strong competitors to traditional statistical models and attracted much attention in the wireless communication domain as a result of the accumulation of massive amounts of cellular traffic data and the advancements in machine learning and AI techniques. The use of AI and machine learning techniques has also allowed for the development of more advanced models that can capture complex relationships in the data and make more accurate predictions. These methods have shown promising results in the wireless communication domain and have received a lot of attention from researchers and industry experts. In the early days of machine learning-based traffic prediction, shallow learning methods such as linear regression were commonly used. These methods are helpful in capturing linear relationships in the data, but they may not be as effective at capturing more complex relationships. In the study by Nie et al.\cite{Nie}, a deep learning-based method was proposed that used the low-pass component of a discrete wavelet transform to capture temporal dependencies in the data. This method showed promising results and demonstrated the potential of deep learning for traffic prediction in cellular networks. Recurrent neural networks built on Long Short-Term Memory (LSTM) units are frequently used to capture the temporal aspect of traffic changes\cite{Huang}. Other deep learning models besides LSTMs, like deep belief networks and Gaussian models, have also been applied to modeling traffic in wireless networks. In addition to temporal dependencies, spatial dependencies in cellular traffic can also be important for making accurate predictions. To capture these dependencies, some studies have proposed hybrid deep learning models that combine different types of deep learning algorithms. For example, in the study by Wang et al.\cite{Wang}, a hybrid model was proposed that used auto-encoders to capture spatial dependencies and LSTM networks to capture temporal dependencies. Another approach that has been proposed is to use only the most correlated neighbors to provide spatiotemporal information rather than using all neighboring traffic information. This can help to reduce the amount of data that needs to be processed and can make the prediction process more efficient. In the study by Qiu et al.\cite{Qiu}, this approach was shown to improve the accuracy of traffic predictions in some cases. In some cases, the spatial distribution of cells may not be regular, which can make it challenging to model the spatial relevancy among them. To overcome this, researchers have used graph neural networks, which are a type of deep learning model that can handle irregular data structures such as graphs. Fang et al.\cite{Fang} used a graph neural network to model the spatial relevancy among cells based on their distances to nearby cell towers. This allowed the network to capture the complex spatial relationships among cells, even in cases where the cell distribution was irregular.

It is well-known that advanced deep learning models, such as LSTM networks and graph neural networks, are typically used for modeling traffic demand on large, high-resolution datasets. However, in many cases, the data available for traffic prediction may be more limited and may consist of short, noisy time series. In such cases, it may not be practical to use complex deep-learning models. Additionally, most of the studies in the literature focus on making short-term forecasts using short-term data collected with a high sampling rate. In our case, there is a need for longer-term estimation at the cell level for network planning. This may require using different modeling techniques or more extended time series data to make more accurate predictions.

\section{Dataset Description and Preliminary Analysis}
The analysis performed in this study was conducted on a data set with a large number of cells collected long-term from networks in various metropolitan areas. The data set includes average downlink volume and average user count data from 23127 cells and is recorded weekly from the beginning of May 2021 to the end of December 2021, leading to 31 samples per cell. The original cellular traffic service had a daily time granularity, but because each cell's data is collected on a different day of the week, it is essential to project data on a common weekly basis. So, before standardizing features, we aggregated the data into one-week periods. In addition, extreme values and values with zero downlink volume are dropped from the data, which seems unrealistic from the perspective of the operators, as they can lead to the instability of the trained model. As the training set, we utilize the first 20 weeks of data, and as the testing set, we use the final 11 weeks. To forecast the data for the upcoming two-time step, we use the traffic information from the previous four-time slots. In order to create a reliable cell-level prediction model in a data set with an excessive number of cells, cells with valid data for all weeks were employed in the study's initial phase. Thus, our final data set consisted of 14841 cells.

The inherent heterogeneity in the data makes it difficult to estimate large-scale mobile traffic accurately. One approach to overcome this challenge is to use data fusion techniques to combine multiple sources of data in order to improve the accuracy of traffic forecasts. The geolocation and points of interest (POIs) of a cell are two examples of external factors that have an impact on cellular traffic volume in addition to temporal elements. For example, a cell located near a popular tourist destination or a major transportation hub is likely to experience higher levels of traffic than a cell in a more remote location. Similarly, the time of day and day of the week can also affect traffic levels, with peak times typically resulting in higher volumes of traffic. Specifically, we collected 33 kinds of POIs, including categories such as residential area, hospital, industrial park, etc. However, in this study, areas of interest covering 95\% of the cells were utilized as is, while the rest were labeled as others, resulting in a total of 12 points of interest. A detailed description and distribution of each category are displayed in Table\ref{tab:table1} and Figure\ref{fig:fig1}. The final representation is formed by the number of cells that fall within each category. The residential regions constitute a significant portion of the cells in the dataset, as seen by the pie chart.

\begin{table}[]
	\centering
	\caption{POIs and corresponding cell numbers}
	\label{tab:table1}
	\begin{tabular}{cc}
		\hline
		\rowcolor[HTML]{C0C0C0} 
		\multicolumn{1}{l}{\cellcolor[HTML]{C0C0C0}Point of Interest (POI)} & \multicolumn{1}{l}{\cellcolor[HTML]{C0C0C0}Number of Cells} \\ \hline
		1-Low Rise Residential Area    & 7484 \\ \hline
		2-High Rise Residential Area   & 2285 \\ \hline
		3-Industrial Park              & 1468 \\ \hline
		4-Colleges and Universities    & 1156 \\ \hline
		5-Others                        & 826  \\ \hline
		6-Village                      & 541  \\ \hline
		7-Office Building              & 360  \\ \hline
		8-Hospital                     & 342  \\ \hline
		9-Commercial Center            & 257  \\ \hline
		10-Enterprises and Institutions & 237  \\ \hline
		11-Urban Road                   & 235  \\ \hline
		12-Square Park                  & 122  \\ \hline
	\end{tabular}
\end{table}

\begin{figure}
	\centering
	\caption{Pie chart of POI distribution.}
	\label{fig:fig1}
	\begin{tikzpicture}
		\pie{
		48.87/1,
		14.92/2,
		9.59/3,
		7.55/4s,
		5.40/5,
		3.53/6,
		2.35/7,
		2.23/8,
		1.68/9,
		1.55/10,
		1.53/11,
		0.80/12
	}
	\end{tikzpicture}
\end{figure}

\section{Methodology}
\begin{figure}
	\centering
	\includegraphics[width=0.5\linewidth]{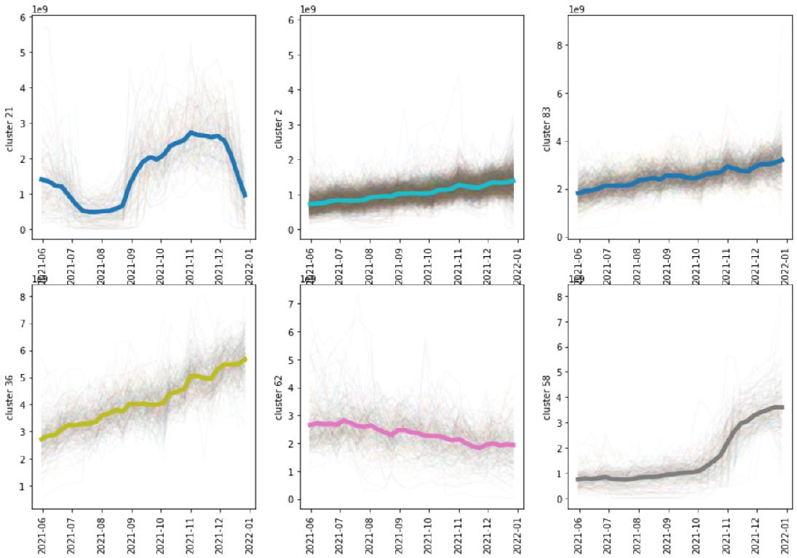}
	\caption{Temporal distributions of cells located in the same region.}
	\label{fig:cluster}
\end{figure}

Estimating large-scale mobile traffic accurately can be challenging due to the inherent heterogeneity in the data. This heterogeneity can be observed in both the spatial and temporal dimensions of the data, with different patterns often being observed in different regions at different times. This makes it difficult to develop models that can accurately forecast traffic in all regions and at all times. Additionally, the large scale of mobile traffic data can make it difficult to process and analyze effectively, which can further complicate the forecasting process. Traffic data show complex spatial dependencies. Even though identical traffic patterns are frequently seen in the same functional areas, sometimes extremely different temporal patterns can be seen even at comparable points of interest. For instance, despite being in the same region, Figure\ref{fig:cluster} clearly shows cells with distinct temporal distributions. Due to aforementioned difficulties, the majority of time series prediction models in use today are unable to perform the task of large-scale traffic forecast directly. It is usual practice to break up the enormous amount of data into smaller subgroups and train each one separately to produce results for the entire cell.

The dataset obtained after the preprocessing applied in this study was first being divided into subgroups according to each point of interest, and then each subgroup is being clustered into 50 groups with the K-Means time series clustering method according to their temporal distribution. This allows for the identification of 50 distinct patterns or trends within each subgroup. By limiting the analysis to clusters that contain at least 20\% of the cells in the cluster with the largest number of cells, the analysis is able to exclude outliers that may not be representative of the broader trend. The K-Means clustering algorithm is a popular method for grouping data into clusters based on their similarity. In the context of time series data, the algorithm can be used to identify patterns and trends in the data and group them into distinct clusters. This approach can help to improve the accuracy of traffic forecasts by allowing for more granular analysis of the data. By taking into account the specific patterns and trends within each subgroup, it is possible to develop more accurate models for predicting traffic volumes.

Once the data was clustered, the proposed N-Beats-based model was trained on each cluster to make predictions for the traffic in that specific point of interest. The N-Beats architecture is a powerful tool for time series prediction, as it is able to capture complex patterns and dependencies in the data\cite{N-BEATS}. Some of the key benefits of the N-Beats architecture include its interpretability, flexibility, and efficiency. Unlike some other deep learning models, N-Beats is designed to be easily understood and interpreted by humans. This means that researchers and practitioners can better understand how the model is making predictions and identify any potential issues or areas for improvement. Another key benefit of N-Beats is its flexibility. The architecture is designed to be applicable without modification to a wide range of target domains, making it a versatile tool for time series prediction. This allows for its use in a variety of applications and scenarios, including traffic forecasting. Finally, N-Beats is also known for its efficiency and speed of training. This makes it a practical and scalable solution for time series prediction.

\begin{figure}
	\centering
	\includegraphics[width=0.7\linewidth]{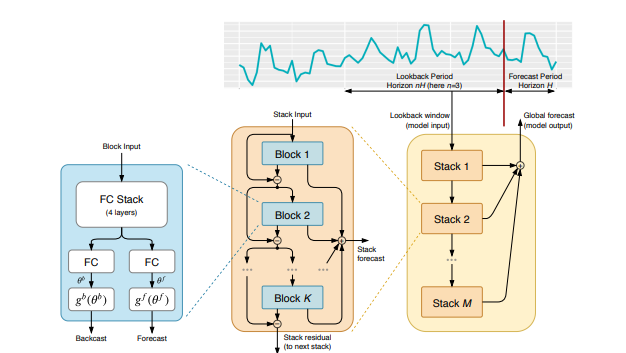}
	\caption{The architecture of the N-Beats Model\cite{N-BEATS}}
	\label{fig:fig2}
\end{figure}

Figure\ref{fig:fig2} shows the architecture of the N-Beats model. The basic building block of N-Beats is a multi-layer fully connected (FC) network with rectified linear unit (ReLU) nonlinearities. This network is used to predict the basis expansion coefficients both forward and backward in time, allowing N-Beats to make forecasts over a wide range of time horizons. The blocks in N-Beats are organized into stacks using a doubly residual stacking principle, which allows the model to build a very deep neural network with interpretable outputs. The forecasts from each block are then aggregated in a hierarchical fashion, which helps improve the overall accuracy of the model's forecasts.

The N-Beats model was proposed for univariate time series estimation. However, in our example, we want to estimate both the average user count and the downlink volume. To enable the simultaneous modeling of several time series, we suggest a multi-branch N-Beats-based architecture. The architecture of the proposed multi-branch model is demonstrated in Figure\ref{fig:fig2}.

\begin{figure}
	\centering
	\includegraphics[width=0.4\linewidth]{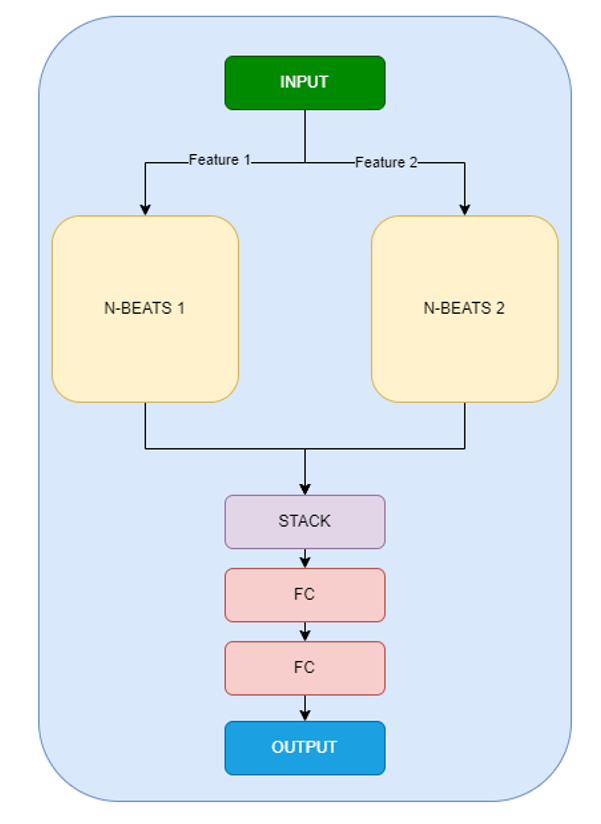}
	\caption{The architecture of the multi-branch model}
	\label{fig:fig3}
\end{figure}

N-Beats uses skip connections in a unique way to improve its ability to make forecasts. The use of skip connections in N-Beats allows the model to make more accurate forecasts by combining information from multiple blocks in a way that is different from traditional neural networks. In N-Beats, skip connections are used to allow subsequent blocks to have easier access to the forecast outputs from previous blocks, which can help improve the overall accuracy of the model's forecasts. By removing the "backcast" outputs from the inputs to subsequent blocks, N-Beats is able to make more accurate forecasts over time. Each block also has its own "forecast" output, which is combined with the forecasts from subsequent blocks to provide a final, combined forecast. 

The length of the input data, or the "lookback period," is a critical parameter in the model. The model uses this data to learn the patterns and trends in the time series. Once the model has learned the behavior of the time series over the lookback period, it uses this information to make predictions for the "forecast period" or the number of future points it is predicting. The length of the forecast period is also a key parameter in the model, as it determines how far into the future the model is able to make predictions.

\section{Experimental Results and Discussion}
In this section, we present the findings from experiments conducted to evaluate performance in order to show how well the suggested model predicts mobile network traffic on a large scale. First, the key parameters and experimental settings are provided. We next compare predictions with ground truth for two randomly chosen cells, as well as the overall prediction performance of the proposed model and naïve baseline method on the test set. 

Various error metrics have been suggested in the literature to assess prediction performance. In the context of this article, the mean absolute percent error (MAPE) is being used to evaluate the performance of the proposed model for predicting mobile network traffic. By focusing on this metric, we are able to provide a clear and understandable measure of the model's performance, which can be easily interpreted by end users.

\begin{table}[]
	\centering
	\caption{Model Performance on each POI}
	\label{tab:results}
	\begin{tabular}{ccc}
		\hline
		\textbf{}                    & \multicolumn{2}{c}{\textbf{MAPE Results}}      \\ \hline
		\textbf{POI}                 & \textbf{Proposed Model} & \textbf{Naive Model} \\ \hline
		Low Rise Residential Area    & 14.88                   & 19.81                \\ \hline
		High Rise Residential Area   & 15.69                   & 20.23                \\ \hline
		Industrial Park              & 16.77                   & 19.56                \\ \hline
		Colleges and Universities    & 165.99                  & 366.48               \\ \hline
		Others                       & 38.15                   & 43.72                \\ \hline
		Village                      & 17.25                   & 19.48                \\ \hline
		Office Building              & 27.72                   & 28.45                \\ \hline
		Hospital                     & 16.97                   & 19.74                \\ \hline
		Commercial Center            & 156.89                  & 282.36               \\ \hline
		Enterprises and Institutions & 34.23                   & 28.75                \\ \hline
		Urban Road                   & 19.92                   & 20.12                \\ \hline
		Square Park                  & 15.43                   & 18.72                \\ \hline
		\textbf{Overall}             & \textbf{19.72}          & \textbf{34.11}       \\ \hline
	\end{tabular}
\end{table}

As can be seen in the Table\ref{tab:results}, the proposed model has defeated the naive model in all areas except Enterprises and Institutions. The results reveal that, despite the fact that the suggested model performs better than the naive model in the commercial center, college, and university zones, the distribution of traffic in these areas varies between the test and training periods.

\begin{figure}
	\centering
	\begin{subfigure}[b]{0.45\textwidth}
		\centering
		\includegraphics[width=\textwidth]{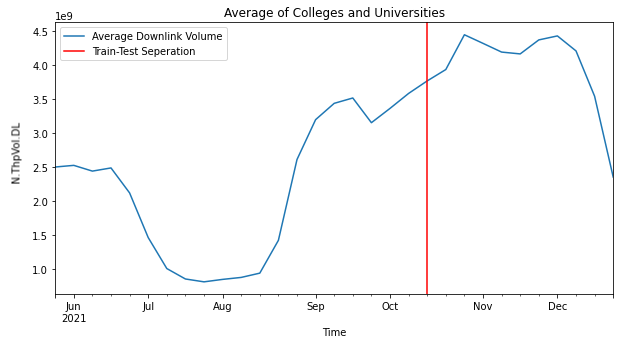}
		\caption{Colleges and Universities}
		\label{fig:cu}
	\end{subfigure}
	\hfill
	\begin{subfigure}[b]{0.45\textwidth}
		\centering
		\includegraphics[width=\textwidth]{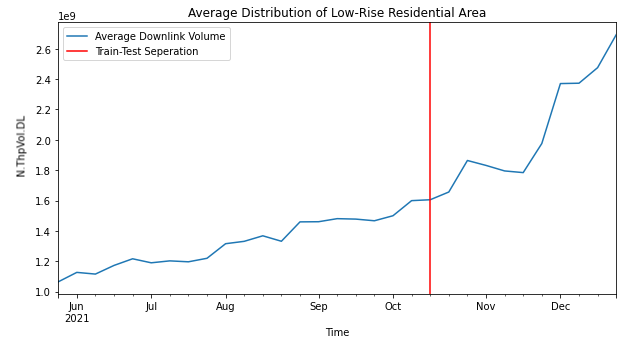}
		\caption{Low Rise Residential Area}
		\label{fig:lr}
	\end{subfigure}
	\caption{Average Time Distribution of the Best and Worst Performing POIs}
	\label{fig:timedist}
\end{figure}

To further analyze the underperforming areas, the Figure\ref{fig:timedist} displays the temporal distributions of the average downlink volumes for the best and worst-performing zones. The results demonstrate a definite trend in the well-performed area, whereas the temporal distribution characteristic of the poor-performance region does not depend on a particular pattern.

\begin{figure}
	\centering
	\includegraphics[width=0.6\textwidth]{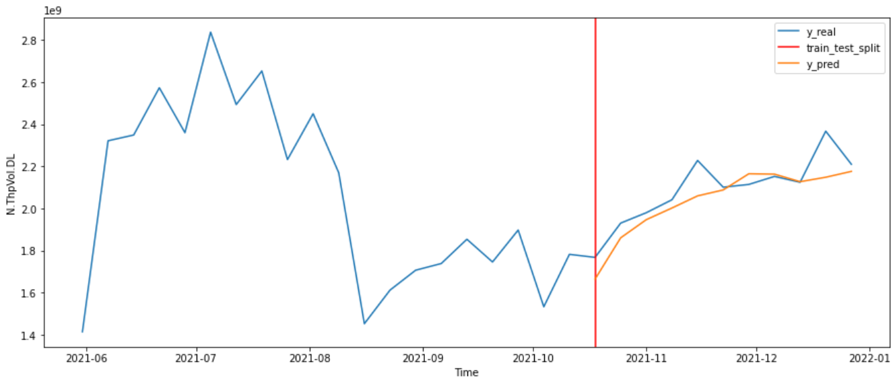}
	\caption{Forecasting result of a sample cell}
	\label{fig:preds}
\end{figure}

It is always more enlightening to illustrate the performance for a specific example in cases where there are too many cells, as presenting results over average values may not inspire the end user. The performance of the proposed model in the test range for a sample cell in a low residential area is shared in Figure\ref{fig:preds}. The red vertical line shows the training and test separation, while the orange line indicates the estimated average downlink volume. The figure indicates that the proposed model is able to accurately predict the average downlink volume for the sample cell, with the estimated values closely tracking the observed values. This suggests that the proposed model is effective at predicting traffic in this particular context and may be useful for forecasting network performance in similar POIs. 

Overall, the results suggest that the proposed model has the potential to be an effective tool for predicting traffic in various areas. Of course, further testing and analysis would be needed to confirm this and determine the model's broader applicability.

\section{Conclusion}
High-precision traffic prediction is crucial for the planning and scheduling of network resources in the dynamic environment of 5G networks, as well as for the dependable and effective transmission of network data. In addition to capturing the pattern diversity and similarity of cellular traffic of various urban functional sectors, this study also presents a method for classifying cells with different temporal distribution characteristics into unique groups. In addition, by utilizing the potent capability of the Neural basis expansion analysis for interpretable time series forecasting architecture, a novel approach for cell-level and long-term traffic prediction from data with a low sampling rate has been provided. The model can accurately estimate 5G traffic, according to the analysis and verification of actual data. The results demonstrate that the method suggested in this study has some advantages over the naive forecasting model, having higher fitting degree, smaller error, and more accurate prediction. This model could definitely use some development, though. For instance, the factors influencing 5G traffic are not fully taken into account. In order to effectively respond to the actual circumstances of 5G use, it is imperative that associated 5G traffic parameters be considered across numerous dimensions in further studies.

\bibliographystyle{unsrt}

\end{document}